# Comparative Analysis of Listwise Reranking with Large Language Models in Limited-Resource Language Contexts

Yanxin Shen*
Simon Fraser University
British Columbia, Canada
yanxin_shen@sfu.ca

Lun Wang
Duke University
North Carolina, USA
lun.wang@alumni.duke.edu

Chuanqi Shi
University of California San Diego
California, USA
chs028@ucsd.edu

Shaoshuai Du
University of Amsterdam
Amsterdam, Netherlands
s.du@uva.nl

Yiyi Tao
Johns Hopkins University
Maryland, USA
ytao23@jhu.edu

Yixian Shen
University of Amsterdam
Amsterdam, Netherlands
y.shen@uva.nl

Hang Zhang
University of California San Diego
California, USA
haz006@ucsd.edu

***Abstract*—Large Language Models (LLMs) have demonstrated significant effectiveness across various NLP tasks, including text ranking. This study assesses the performance of large language models (LLMs) in listwise reranking for limited-resource African languages. We compare proprietary models RankGPT3.5, Rank4o-mini, RankGPTo1-mini and RankClaude-sonnet in cross-lingual contexts. Results indicate that these LLMs significantly outperform traditional baseline methods such as BM25-DT in most evaluation metrics, particularly in nDCG@10 and MRR@100. These findings highlight the potential of LLMs in enhancing reranking tasks for low-resource languages and offer insights into cost-effective solutions.**

## I. Introduction

Researchers have explored the utility of LLMs for text ranking, both as retrievers and as rerankers using listwise or pointwise methods [1]. This makes it crucial to assess the potential of advanced models like LLMs [7]. Listwise approaches are particularly advantageous due to the large context size that LLMs can handle. These models consider multiple documents to generate a ranked list. Ma et al. (2023b) also indicated that LLMs can generalize listwise reranking effectively [3].

Our research is primarily driven by these questions:

- How effective are LLMs at performing listwise reranking for languages with limited resources?
- How do proprietary LLMs such as RankGPT3.5, Rank4o-mini, RankGPTo1-mini, and RankClaude-sonnet compare to traditional ranking methods like BM25-DT in cross-lingual settings?
- What are the specific challenges of reranking passages translated from low-resource African languages, and how well do LLMs handle these challenges?

In this study, we explore these questions by evaluating the performance of LLMs in reranking tasks where the queries are in English, but the documents are translated from African languages with limited resources. African languages, such as Hausa, Somali, Swahili, and Yoruba, represent a unique challenge in NLP due to the scarcity of labeled data and linguistic resources. This makes it crucial to assess the potential of advanced models like LLMs, which have demonstrated impressive generalization abilities across a wide range of NLP tasks.

The motivation behind our research stems from the rapid advancements in LLMs and their applications in information retrieval (IR). Traditional methods, such as BM25, rely heavily on term-frequency-based mechanisms and often struggle in low-resource languages where vocabulary and syntax vary significantly from English. LLMs, on the other hand, leverage pre-trained knowledge and are capable of capturing semantic relationships, even in cross-lingual scenarios. This opens up new possibilities for improving the quality of search and retrieval tasks in underrepresented languages.

This study seeks to address these questions by evaluating the performance of RankGPT3.5, RankGPT4o-mini, RankGPTo1-mini and RankClaude-sonnet in cross-lingual retrieval contexts. We employ CIRAL to create cross-lingual retrieval scenarios via query or document translation.

To thoroughly evaluate these models, we designed experiments to measure their reranking performance on these cross-lingual tasks. Specifically, African language passages were ranked based on their relevance to English queries. The reranking effectiveness was assessed using two primary evaluation metrics: Normalized Discounted Cumulative Gain at rank 10 (nDCG@10) and Mean Reciprocal Rank at rank 100 (MRR@100). These metrics offer a comprehensive view of the models' precision and overall ranking quality.

Our findings indicate that cross-lingual reranking with RankGPT3.5, RankGPT4o-mini, RankGPTo1-mini, and RankClaude-sonnet consistently outperforms traditional baseline methods like BM25-DT for low-resource languages. Among these models, RankGPTo1-mini demonstrates the best overall performance, especially for languages like Hausa, Swahili and Yoruba. RankGPT3.5, RankGPT4o-mini and RankClaude-sonne also perform well, achieving significant improvements in reranking tasks. These results underscore the effectiveness of these LLMs in enhancing retrieval tasks in cross-lingual settings, especially for limited-resource African languages.

In the following sections, we will discuss related work, the experimental setup, the results of our experiments, and a detailed analysis of the performance differences between the models. Finally, we will provide insights into the potential future directions for improving cross-lingual reranking in low-resource languages.

## II. Background and Related Work

Information retrieval (IR) involves searching a corpus of documents [9] for a query with the goal of retrieving the top relevant documents. Contemporary IR systems typically employ a multi-stage process where a retriever generates a list of candidate documents. Earlier approaches used models such as TFIDF or BM25 [4] as the initial stage retrievers. However, the advent of pretrained text encoders like BERT, which provide more advanced dense representations, has led to increased research and adoption of these improved methods.

Recently, there has been an increased focus on the performance and efficiency of Transformer decoders within multistage information retrieval (IR) systems [1]. Researchers have explored the fine-tuning of GPT-like models [8].

While GPT-3 outperforms BM25 for these languages, it still does not meet the performance levels of supervised reranking benchmarks. Cross-lingual Information Retrieval (CLIR) addresses the challenge of retrieving documents. In this research, we investigate document translation approaches, as the primary aim is to assess the performance of LLMs as listwise rerankers.

With the continuous advancement of machine learning algorithms and the availability of massive data, significant progress has been made in the field of natural language processing (NLP). Among these developments, GPT-3.5-Turbo represents a major breakthrough in NLP models. As a generative language model, GPT-3 can predict the next word in a given sentence based on context, thereby generating human-like text. Currently, it is one of the most advanced language models available, featuring 175 billion parameters. GPT-3.5-Turbo builds upon GPT-3 by adding additional functionalities and parameters, making it more powerful and capable of achieving higher accuracy and better performance across various NLP tasks.

GPT-3.5 is being replaced by GPT-4o Mini in ChatGPT. In recent years there have been significant advances in AI intelligence and cost reductions. The cost per token for GPT-4o Mini has decreased by 99% when compared to the text-davinci003, which is a less capable version from 2022.

OpenAI o1 is a large language model trained using reinforcement learning, focusing on enhancing complex reasoning abilities. It "thinks first" before answering questions, generating a longer chain of reasoning before providing an answer, which makes its reasoning more accurate and in-depth. It outperformed previous models in multiple tasks and showed significant improvements in human preference evaluations and logical reasoning. The model is trained using reinforcement learning, teaching it how to use reasoning chains in complex problems. This enables the model to excel at tasks that require step-by-step reasoning, allowing it to gradually adjust its

| | **Source** | | **nDCG@10** | | | | **MRR@100** | | | |
|---|---|---|---|---|---|---|---|---|---|---|
| | **Prev.** | **Top-k** | **ha** | **so** | **sw** | **yo** | **ha** | **so** | **sw** | **yo** |
| (1) BM25-DT | None | $\|C\|$ | 0.0992 | 0.1358 | 0.2026 | 0.3260 | 0.1340 | 0.2717 | 0.3180 | 0.4191 |
| *Cross-lingual Reranking: English queries, passages translated to English from African languages* | | | | | | | | | | |
| (2a) RankGPT3.5 | BM25-DT | 100 | 0.1388 | **0.2708** | 0.1957 | 0.2771 | 0.2391 | **0.4828** | 0.2549 | 0.4381 |
| (2b) RankGPT4o-mini | BM25-DT | 100 | 0.1965 | 0.2270 | **0.2282** | 0.3486 | 0.3074 | 0.3823 | 0.3480 | 0.5113 |
| (2c) RankGPTo1-mini | BM25-DT | 100 | **0.2074** | 0.1947 | 0.2135 | **0.3703** | 0.3047 | 0.3210 | **0.3636** | **0.5530** |
| (2d) RankClaude-sonnet | BM25-DT | 100 | 0.2027 | 0.2007 | 0.2111 | 0.2820 | **0.3366** | 0.3761 | 0.2960 | 0.4011 |

Table 1 provides a comparative analysis of reranking results in cross-lingual contexts. In the cross-lingual setup, English queries from CIRAL are paired with passages translated to English from African languages.

strategy and correct mistakes. The performance of o1 will continue to improve as training and usage time increase. This accumulation of the reinforcement learning process allows it to perform better and better on complex tasks.

o1 has outperformed GPT-4.0 in multiple reasoning tasks, particularly in those requiring complex logic and multi-step reasoning, such as mathematics and programming competition problems (AIME, Codeforces, GPOA). Its performance on advanced mathematics, science, and physics problems exceeds that of a PhD level.

Claude-3-5-sonnet is another advanced language model developed by Anthropic, a prominent player in the AI landscape. This model has demonstrated impressive capabilities in various NLP tasks, showcasing its potential to compete with OpenAI's offerings. Claude-3-5-sonnet is designed to be more efficient than GPT-3.5 while maintaining high levels of accuracy and coherence. Its strength lies in generating creative and informative text, making it suitable for applications such as content creation and storytelling.

One of the key advantages of Claude-3-5-sonnet is its ability to understand and respond to complex prompts and questions. It can generate detailed and informative responses, even when presented with ambiguous or incomplete information. This makes it a valuable tool for tasks that require deep understanding and analysis of text.

## III. Method

### A. Listwise Reranking

Large Language Models (LLMs) utilize listwise reranking to evaluate and rank the relevance of various documents in response to a single prompt. This method has demonstrated greater efficiency compared to pointwise or pairwise reranking techniques. Our approach exclusively employ listwise reranking.

### B. Prompt Design

We adopt RankGPT (Sun et. al., 2023) as modified by Pradeep et. al. (2023a). The input prompts and generated completion are:

Input Prompt:
SYSTEM
You are RankGPT, an intelligent assistant
that can rank passages based on their relevancy
to the query.
USER
I will provide you with {num} passages,
each indicated by number identifier [].
Rank the passages based on their relevance
to the query: {query}.
[1] {passage 1}
[2] {passage 2}
...
[num] {passage num}
Search Query: {query}
Rank the {num} passages above based
on their relevance to the search query.
The passages should be listed in descending
order using identifiers. The most relevant
passages should be listed first. The output
format should be [] > [], e.g., [1] > [2].
Only respond with the ranking results, do not
say any word or explain.
Model Completion:
[10] > [4] > [5] > [6] ... [12]

### C. LLM Zero-Shot Translations

Document translation serves as a method to address language barriers encountered when retrieving and reranking data across different languages. In this context, we assess the performance of Large Language Models (LLMs). We create zero-shot translations from African language into a certain language for a specific LLM and subsequently apply reranking based on these translations. This method enables us to assess how well an LLM performs in African languages and explore the relationship between the quality of translations and the reranking outcomes. Our prompt:

Input Prompt:

Documents: {doc }

Translate this doc from {African language} to {certain language}.

Only return the translation, don't say any other word.

Model Completion:

{Translated query}

## IV. Experimental Setup

### A. Models

We employ four models to conduct zero-shot reranking. These include proprietary reranking LLMs such as RankGPT3.5, which utilizes the gpt-3.5-turbo model; RankGPT4o-mini, which employs the gpt 4o-mini model; and RankGPTo1-mini, based on the OpenAI o1-mini model, all leveraging OpenAI's API [6]. We use RankClaude-sonnet, based on the Claude-3-5-sonnet, a reranking LLM developed from Anthropic [5] to deliver competitive results comparable to those of the OpenAI models.

### B. Test Collection

Models are assessed using CIRA, a CLIR test suite comprising four African languages: Hausa, Somali, Swahili, and Yoruba [3]. This collection features natural language factoids in English, with corresponding passages. For our document translation scenario, we can also utilize CIRAL's translated-passage collection. We report results based on nDCG@10 and MRR@100, adhering to the standards of the test collection.

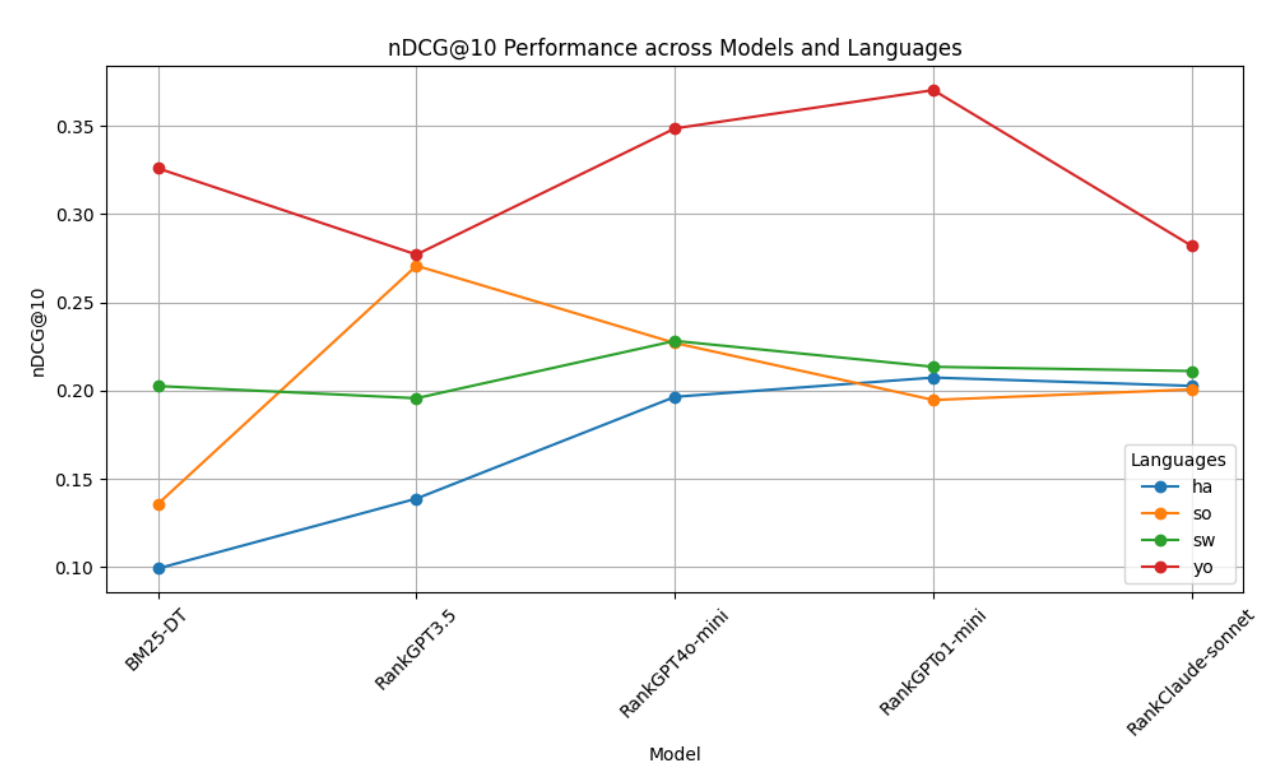

Figure 1: nDCG@10 Performance across Models and Languages

### C. *Configurations*

The open-source Pyserini toolkit is employed for retrieval using BM25 as the initial phase [4]. In this process, the default English tokenizer is used for translated passages, while whitespace tokenization is applied to native languages if needed. We examine retrieval performance in the first stage by utilizing document translation (BM25 DT).

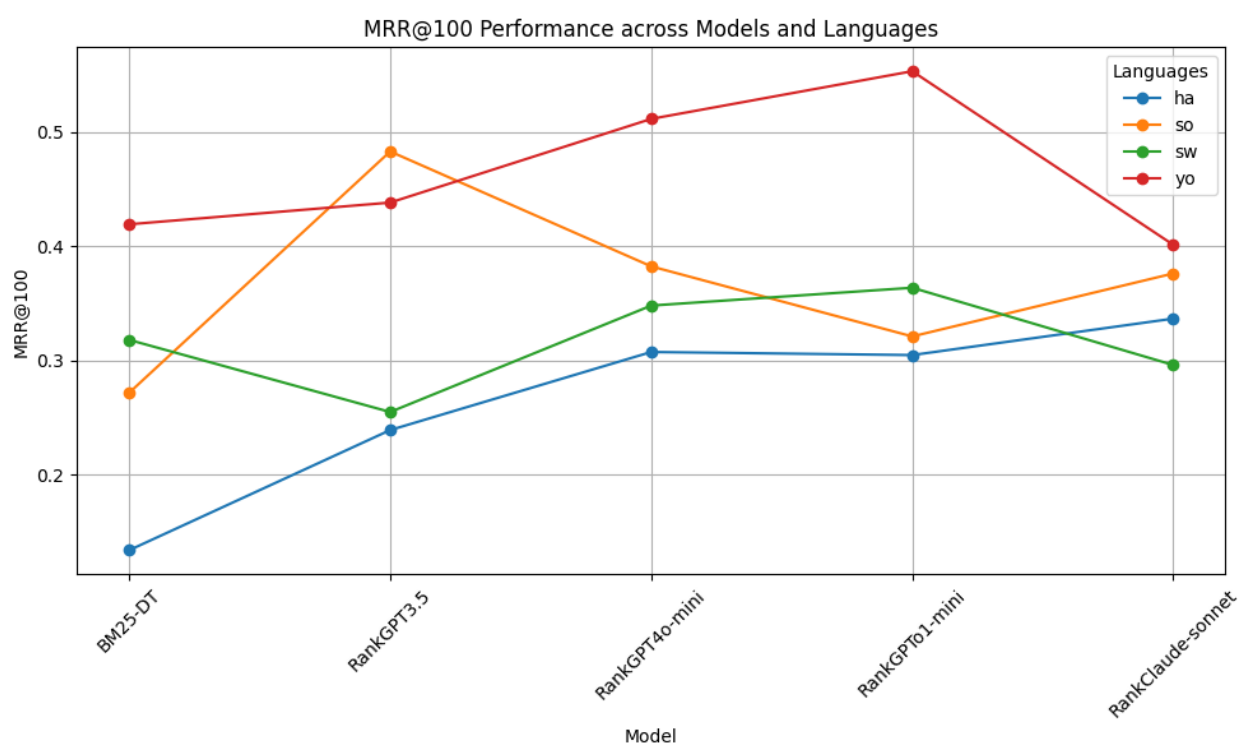

Figure 2: MRR@100 Performance across Models and Languages

For context size, RankGPT3.5 has a context size of 4096 tokens, RankGPT-4o-mini utilizes 16384 tokens, O1 mini uses 65536 tokens, while RankClaude-3-5-sonnet has a context size of 8192 tokens. Due to the limitations of the context window, BM25 retrieves only the top 100 most relevant documents for each query. This ensures that the model operates within its capacity while maximizing the relevance of the information presented.

## V. Results

Table 1 illustrates outcomes of cross-lingual reranking from CIRAL's queries, passages, and reranking scenarios in English. In Row (1), we display the scores for our baseline method. Row (2) details the cross-lingual reranking results for various LLMs, with BM25 DT employed for retrieval in the initial stage due to its nice effectiveness. LLMs are anticipated to perform better with English translations due to their greater familiarity with English, despite their multilingual capabilities.

Row (2a) through Row (2d) present the performance of proprietary LLMs, including RankGPT3.5, RankGPT4o-mini, RankGPTo1-mini, and RankClaude-sonnet, respectively. The reranking results highlight significant improvements over the baseline, especially in languages like Hausa, Yoruba and Somali, where the LLMs consistently outperform BM25 DT in both nDCG@10 and MRR@100 metrics. Notably, RankGPTo1-mini achieves the highest scores across multiple languages, indicating its strong cross-lingual capabilities.

These results suggest that LLMs can capture complex semantic relationships better than traditional retrieval models in cross-lingual scenarios, making them more effective for languages with limited resources. Additionally, the use of BM25 DT as the initial retrieval method allows the LLMs to refine the candidate passages further, leveraging their advanced contextual understanding. Fig. 1 and Fig. 2 show nDCG@10 and MRR@100 performance across models and languages.

The analysis of these models also reveals that while all LLMs outperform BM25 DT, the degree of improvement varies depending on the language. For example, Yoruba consistently benefits the most from reranking, while Hausa shows more modest gains. This indicates that some languages may still pose challenges for even advanced models, possibly due to the unique syntactic or semantic characteristics of these languages or the availability of training data.

## VI. Conclusion

In this paper, we evaluate the performance of LLMs in reranking tasks involving limited-resource languages. Additionally, it's possible that the reranking performance of LLMs in African language scenarios benefits from high-quality translations, which we will demonstrate further.

We compared the performance of RankGPT3.5 with RankGPT4o-mini, RankGPTo1-mini and RankClaude-sonnet. RankGPTo1-mini emerged as the most effective reranker for tasks involving limited-resource languages. Particularly, RankGPT-4o-mini, RankGPTo1-mini and RankClaude-sonnet not only significantly improves reranking performance but also offers substantial cost savings [2], making it a highly efficient solution, compared with RankGPT4. Due to the increasing effectiveness of open-source LLMs in this area, they hold great potential for enhancing tasks involving low-resource languages..

We believe our work demonstrates the potential of large-scale language models for handling tasks involving limited-resource languages and showcases the promising future for these languages. Additionally, we hope this research will inspire further efforts to enhance the effectiveness of LLMs in addressing challenges related to limited-resource languages.